\documentclass[journal]{IEEEtran}

\usepackage{times}
\usepackage{epsfig}
\usepackage{graphicx}
\usepackage{amsmath}
\usepackage{amssymb}
\usepackage{bbm}
\usepackage{cite}
\usepackage{float}
\usepackage{algorithm}
\usepackage{algorithmicx}
\usepackage{algpseudocode}
\usepackage{mathtools}
\usepackage{url}

\usepackage[table]{xcolor} 
\usepackage[para,online,flushleft]{threeparttable}
\usepackage{booktabs}
\usepackage{longtable}
\usepackage{tablefootnote}
\usepackage{etoolbox}
\appto\TPTnoteSettings{\footnotesize}

\usepackage{arydshln}
\usepackage{caption}
\usepackage{subcaption}
\usepackage{makecell}
\usepackage{pifont}
\usepackage{multirow}
\usepackage{booktabs}
\usepackage{chngcntr}
\usepackage{float}
\usepackage{diagbox}

\usepackage{amsmath}
\usepackage{romannum}

\makeatletter
\newcommand\threepart@subtable{
  \caption@setoptions{threepartsubtable}%
  \caption@ORI@threeparttable
}

\makeatother

\DeclareMathOperator*{\argmin}{arg\,min}

\makeatletter
\let\NAT@parse\undefined
\makeatother
\usepackage[breaklinks=true,bookmarks=false]{hyperref}
\hypersetup{
    colorlinks=true,
    linkcolor=blue,
    filecolor=magenta,
    urlcolor=cyan,
}

\title{Take-over Time Prediction for Autonomous Driving in the Real-World: Robust Models, Data Augmentation, and Evaluation}

\author{Akshay Rangesh$^{\dag}$, Nachiket Deo$^{\dag}$, Ross Greer$^{\dag}$, Pujitha Gunaratne and Mohan M. Trivedi

\thanks{
$^\dag$authors contributed equally.
}%
\thanks{Akshay Rangesh, Nachiket Deo, Ross Greer and Mohan M. Trivedi are with the Laboratory for Intelligent and Safe Automobiles, UC San Diego.}%
\thanks{Pujitha Gunaratne is with Toyota Collaborative Safety Research Center.}%
\thanks{This work has been submitted to the IEEE for possible publication. Copyright may be transferred without notice, after which this version may no longer be accessible.}

}

\begin{document}

\maketitle

\begin{abstract}
Understanding occupant-vehicle interactions by modeling control transitions is important to ensure safe approaches to passenger vehicle automation. Models which contain contextual, semantically meaningful representations of driver states can be used to determine the appropriate timing and conditions for transfer of control between driver and vehicle. However, such models rely on real-world control take-over data from drivers engaged in distracting activities, which is costly to collect. Here, we introduce a scheme for data augmentation for such a dataset. Using the augmented dataset, we develop and train take-over time (TOT) models that operate sequentially on mid and high-level features produced by computer vision algorithms operating on different driver-facing camera views, showing models trained on the augmented dataset to outperform the initial dataset. The demonstrated model features encode different aspects of the driver state, pertaining to the face, hands, foot and upper body of the driver. We perform ablative experiments on feature combinations as well as model architectures, showing that a TOT model supported by augmented data can be used to produce continuous estimates of take-over times without delay, suitable for complex real-world scenarios.
\end{abstract}


\section{Introduction}
\begin{figure}[t]
    \center{\includegraphics[width=0.85\linewidth]
    {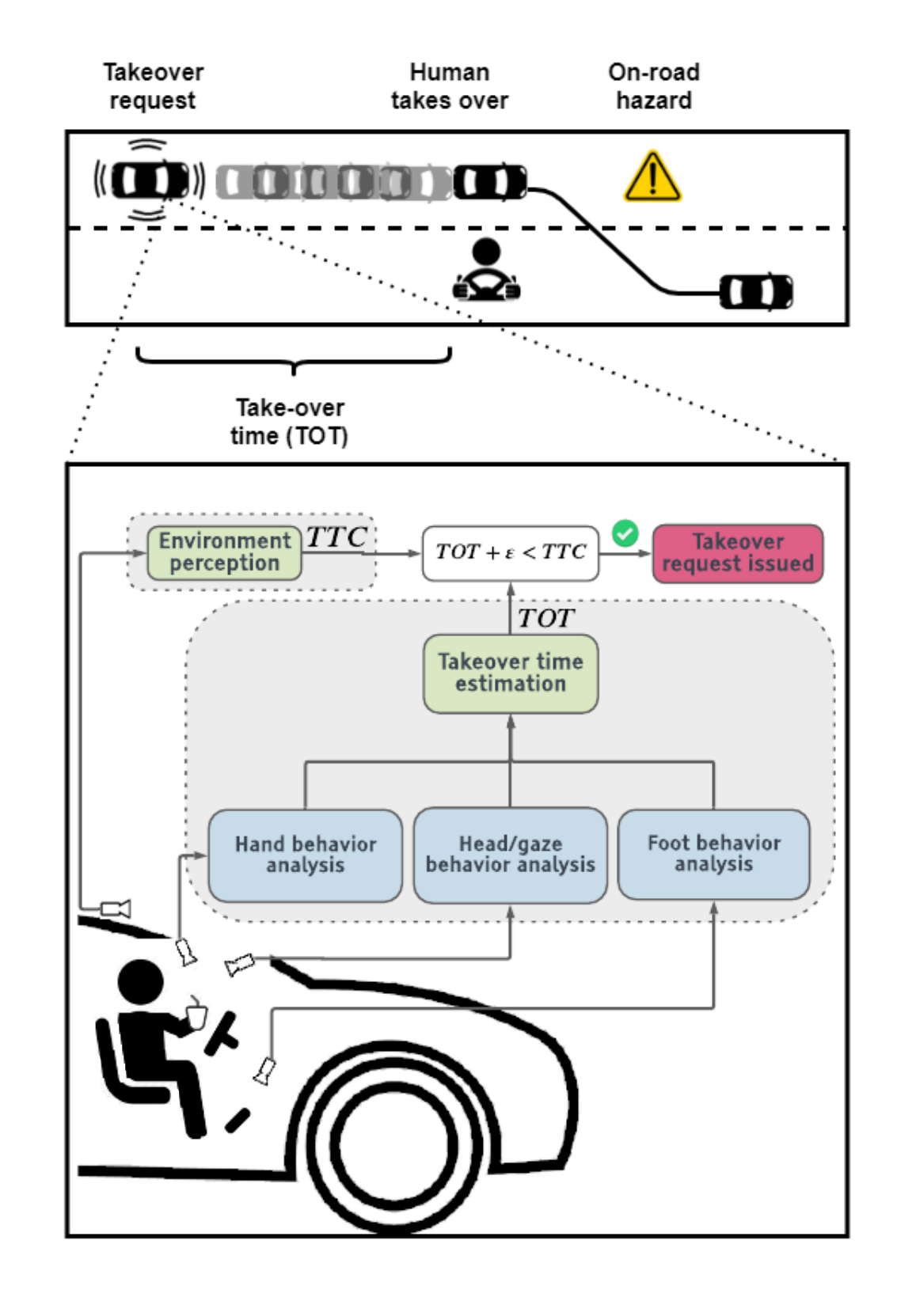}}
    \caption{\textbf{Role of take-over time (TOT) prediction:} We propose a model for predicting TOT during control transitions based on driver behavior. The proposed model can be used in conjunction with time-to-collision estimation to determine whether to issue a take-over request and transfer control to the human, or to deploy active safety measures for collision avoidance.}
  \label{fig:motivation}
\end{figure}


\IEEEPARstart{M}otivations for studying driver behavior in highly automated vehicles can be found aplenty in human factors studies (e.g. \cite{kircher2017minimum}, \cite{jensen2010studying}, \cite{tice2021driver}, \cite{gaspar2019effect}, \cite{benedetto2011driver}, \cite{ojstervsek2019eye}). It is widely regarded that as soon as the level of cognitive stimulation falls below a person's own comfortable ``set point", the person will seek out alternate/additional sources of information, leading to distraction (e.g. \cite{isaza2019dynamic}, \cite{eastwood2012unengaged}, \cite{charlton2011driving}, \cite{bernstein2015texting}). This makes the intermediate levels of automation (as per NHTSA \cite{nhtsa} or SAE \cite{williams2021automated}) very dangerous, causing problems such as inattention, trust, skill atrophy, complacency, etc.~\cite{casner2016challenges}. The authors in ~\cite{casner2016challenges} postulate that rising levels of automation will lead to declining levels of awareness. They also state that most problems are expected to arise in systems that take the driver out of the loop, yet these are the very systems that drivers want, because they free the driver to do something else of interest. Elsewhere, the authors in ~\cite{kyriakidis2017human} emphasize the \textit{irony of automation}, whereby ``the more advanced a control system is, the more crucial may be the contribution of the human operator". They also acknowledge that decades of research has shown that humans are not particularly good at tasks that require vigilance and sustained attention over long periods of time \cite{warm2008vigilance}. 

All above points seem to suggest that a drop in attention is inherent in human behavior. Thus, it is not a matter of \textit{if}, but \textit{when} the driver will resort to non-ideal behavior. This makes the safe and smooth handling of control transitions, which entail the transfer of vehicle controls from the autonomous agent to the human driver and vice versa, extremely important and timely. Consider the scenario illustrated in Fig.~\ref{fig:motivation}, indicating the transition of control from an autonomous agent to the human driver to be a function of the driver state. We propose that a system that takes the state of the driver into account can decide between handing over control if the driver is ready, versus coming to a safe and smooth halt if not. Driver state can also dictate how and when a takeover alert must be supplied to ensure an uneventful transition of control.

In this paper, we focus on transitions from the autonomous agent to the human driver. In particular, we consider scenarios where 
limits of the autonomous system are reached. For example, an unforeseen on-road hazard may be detected that needs to be evaded. The conditions for L3 autonomy may be coming to an end with the vehicle leaving a geofenced area, or a traffic jam assist system encountering dissipation of the traffic jam. Such scenarios require timely human intervention within a predictable time window. 
In describing such control transitions, we make use of the take-over time (TOT) metric, defined as the interval of time between a take-over request (TOR) being issued and the assuming of human control. The take-over request could be an auditory/visual/tactile cue used to indicate to the driver that their intervention is immediately needed. Due to the complexity of human attention, we define the assumption of control as the completion of the following three behaviors:
\begin{enumerate}
    \item \textbf{Hands-on-wheel:} hand(s) return to the vehicle's steering control.
    \item \textbf{Foot-on-pedal:} foot returns (from floorboard or hovering) to make contact with any driving pedal.
    \item \textbf{Eyes-on-road:} gaze is directed forward, toward the active driving scene. 
\end{enumerate}
We work with the assumption that these three cues occurring simultaneously are \textit{necessary} to consider the driver both attentive to the scene and in control of the vehicle. We do
note that the three cues may not be \textit{sufficient} to consider the driver attentive and in control. This would additionally depend on factors such as the driver's situational awareness and the corrective/stabilizing maneuver performed post TOR. We limit the scope of this work to predicting the time taken for the above three cues, as a first step towards analysis of control transitions using real-world autonomous driving data. Analysis of situational awareness and corrective maneuvers will be addressed in future work.

As depicted in Fig.~\ref{fig:motivation}, the transition of control from an autonomous agent to the human driver should be a function of both the surrounding scene and the state of the driver. The surrounding scene can be concisely expressed using a metric such as time-to-collision (TTC), whereas the state of the driver can be captured by the predicted TOT. Combined, this forms a criterion for safe control transitions: 
\begin{equation}
    TOT + \varepsilon < TTC,
\end{equation}
where $\varepsilon$ is a marginal allowance that represents the time it takes for the human driver to gain situational awareness and perform a corrective maneuver. A system that takes the state of the driver into account can decide between handing over control if the driver is ready, versus coming to a safe and smooth halt if not. While there are many approaches to accurately estimate TTC, TOT prediction (especially in the real world) remains relatively unexplored. In this paper, we present a long short-term memory (LSTM) model for predicting TOT based on driver behavior prior to the TOR. We train and evaluate our model using a real world dataset of control transitions captured using a commercially available conditionally autonomous vehicle. This work is an extension of our prior work \cite{rangesh2021autonomous}, with three new contributions:
\begin{enumerate}
    \item \textbf{TOT prediction with limited real-world data:} Capturing real-world takeover events in autonomous vehicles is expensive and time-consuming. Thus generating a large enough dataset for training machine learning models can be a challenge. To address this, we propose a data-augmentation scheme to increase the number of training samples by an order of magnitude. Additionally we use transfer learning, and pre-train our TOT prediction models to estimate the driver's observable take-over readiness index (ORI) \cite{deo2019looking}.  
    \item \textbf{Multimodal TOT prediction:} There is inherent uncertainty in predicting the future. The driver could perform multiple plausible sequences of actions after the issued TOR. To model this, we extend the model proposed in \cite{rangesh2021autonomous} to output a multimodal distribution over TOT.
    \item \textbf{Extensive evaluation:} We present a more extensive set of ablation experiments, particularly focused on the above two contributions. We also present additional qualitative analysis of TOT estimates beyond \cite{rangesh2021autonomous}.
\end{enumerate}


\section{Related Research}

\subsection{Vision based driver behavior analysis}
A large body of literature has addressed driver behavior analysis using in-cabin vision sensors. The most commonly addressed task is driver gaze estimation \cite{murphy2008hyhope, tawari2014robust,tawari2014driver,lee2011real,vasli2016driver,fridman2016driver,fridman2016owl,vora2017generalizing,vora2018driver, Naqvi_NIR_2018,Jha_2018,rangesh2020driver}, since the driver's gaze closely relates to their attention to driving and non-driving tasks. Early works relied on head pose estimation \cite{murphy2008hyhope, lee2011real,tawari2014robust, tawari2014continuous} or a combination of head and eye features \cite{doshi2012head, tawari2014driver,vasli2016driver,fridman2016driver,fridman2016owl} for estimating the driver's gaze. More recent work \cite{vora2018driver, vora2017generalizing, Naqvi_NIR_2018,Jha_2018, rangesh2020driver} uses convolutional neural networks (CNNs) to directly map regions around the driver's eyes to gaze zones. In this work, we use the CNN model proposed by Vora \textit{et al.} \cite{vora2018driver} driver gaze analysis. 

Driver hand and foot activity has also been the subject of prior work, being useful cues to gauge the driver's motor readiness. Several approaches have been proposed for detection, tracking and gesture analysis of the driver's hands \cite{das2015performance, ohn2013vehicle,ohn2014beyond,ohn2014hand,borghi2018hands,rangesh2016hidden,deo2016vehicle,molchanov2015hand} using in vehicle cameras and depth sensors. Recently proposed CNN models \cite{yuen2019looking, rangesh2018handynet} accurately localize the driver's hands in image co-ordinates and in 3-D respectively, and further classify hand-activity and held objects. We build upon the model proposed by Yuen \textit{et al.} \cite{yuen2019looking} in this work, for driver hand analysis. Relatively few works have addressed the driver's foot activity \cite{tran2011pedal,tran2012modeling, rangesh2019foot}. However, we believe this is a significant cue for TOT estimation, especially since we estimate the foot-on-pedal time after the TOR. We use the model proposed by Rangesh \textit{et al.} \cite{rangesh2019forced} for driver foot activity analysis.  

There has also been significant research that builds upon cues from driver gaze, hand and foot analysis for making higher level inferences such as driver activity recognition \cite{ohn2014head,braunagel2015driver,behera2018context, roitberg2020cnn, roitberg2020open, martin2019drive, reiss2020deep, behera2020deep}, driver intent or behavior prediction \cite{jain2015car,jain2016recurrent,martin2018dynamics,ohn2014predicting, doshi2008comparative, shia2014semiautonomous, driggs2015improved} and driver distraction detection \cite{liu2016driver,liang2007real,liang2014hybrid,bergasa2006real,li2015predicting,wollmer2011online}. Of particular interest is recent work \cite{deo2019looking}, where the authors map driver gaze, hand and foot activity to the driver's observable take-over readiness index (ORI) obtained via subjective ratings assigned by multiple human observers. We use ORI estimation as a transfer learning task for pre-training our TOT prediction model.

\subsection{Take-over time analysis in autonomous driving}

Take-over time in partial and conditionally autonomous vehicles has been the subject of several recent studies \cite{gold2013take, mok2015timing, radlmayr2014traffic, gold2016taking, korber2016influence, clark2017age, petermeijer2017take, huang2019multimodal, dogan2017transition, eriksson2017takeover, naujoks2019noncritical}. The primary focus of these studies has been to analyze the effect of various human and environmental factors on take-over time and quality. The independent variables analyzed for their effect on TOT are as follows:

\vspace{1mm}
\noindent \textbf{TOT budget (or time to collision):} This corresponds to the time window between the TOR and the imminent collision or system boundary. Gold \textit{et al.} \cite{gold2013take} compare TOT and take-over quality for two different TOT budgets of 5s and 7s. They report longer TOTs for the 7s budget but better take-over quality. Mok \textit{et al.} \cite{mok2015timing} report a similar finding while comparing TOT budgets of 2s, 5s, and 8s, with the 2s case corresponding to significantly worse take-over quality and collision rates.

\vspace{1mm}    
\noindent \textbf{Traffic density:} Radlmayr \textit{et al.} \cite{radlmayr2014traffic} and Gold \textit{et al.} \cite{gold2016taking} analyze the effect of traffic density on TOT and take-over quality, with both studies reporting longer TOTs and worse take-over quality in situations involving high traffic density.

\vspace{1mm}    
\noindent \textbf{Driver age:} Korber \textit{et al.} \cite{korber2016influence} and Clark and Feng \cite{clark2017age} analyze the effect of driver age on TOT by comparing a group of young drivers with a group of old drivers. Korber \textit{et al.} \cite{korber2016influence} report similar TOTs, but different modus operandi -- older drivers brake harder and more often leading to higher TTC. Clark and Feng \cite{clark2017age} report lower TOTs for the young group for a TOT budget of 4.5s, and lower TOTs for the old group for a 7.5s TOT budget.

\vspace{1mm}    
\noindent \textbf{TOR modality:} Petermeijer \textit{et al.} \cite{petermeijer2017take} and Huang \textit{et al.} \cite{huang2019multimodal} compare different modalities for issuing the TOR. Auditory and tactile TORs are considered in \cite{petermeijer2017take} while auditory, tactile and visual TORs and their combinations are considered in \cite{huang2019multimodal}. Both studies report the lowest TOTs for multimodal TORs. Dogan \textit{et al.} \cite{dogan2017transition} analyze the effect of providing the driver anticipatory information about the vehicle and traffic state prior to the TOR, but report similar TOTs with and without the anticipatory information.

\vspace{1mm}    
\noindent \textbf{Non-driving-related tasks (NDRTs):} Several prior works \cite{eriksson2017takeover, radlmayr2014traffic, dogan2017transition, naujoks2019noncritical} have consistently reported worse take-over times or take-over quality when the driver is engaged in a NDRT prior to the take-over, whether the NDRT places visual, cognitive or motor-control based demand on the driver. In this paper, we thus primarily focus on the effect of driver behavior and NDRTs on TOT. In particular, we map the observed NDRTs to feature descriptors of driver gaze, hand and foot activity using vision based models for driver behavior analysis and predict TOT based on these feature descriptors.

\subsection{Take-over time prediction for autonomous driving}
While the studies described in the previous section analyze take-over times under various experimental conditions,
closest to our work are recently proposed machine learning models \cite{ braunagel2017ready, lotz2018predicting, berghofer2019prediction, du2020predicting, hwang2020predicting, pakdamanian2021deeptake} that \textit{predict} TOT prior to the control transition. 

Braunagel \textit{et al.} \cite{braunagel2017ready} and Du \textit{et al.} \cite{du2020predicting} propose binary classifiers that output whether or not the driver is ready to take-over. Gaze activity, NDRT label and a label for situation complexity are used as input features in \cite{braunagel2017ready}, while gaze activity, heart rate variability, galvanic skin response, traffic density and TOT budget are used as inputs in \cite{du2020predicting}. 
Pakdamanian \textit{et al.} \cite{pakdamanian2021deeptake} propose a three class classifier over TOT intervals based on driver gaze activity, heart rate variability, galvanic skin response, NDRT label and vehicle signals. Lotz and Weissenberger \cite{lotz2018predicting} compare various classifiers over 4 TOT intervals trained using features capturing driver's head orientation and gaze activity, along with TTC. Hwang \textit{et al.} \cite{hwang2020predicting} propose a regression model based on hidden Markov models that outputs TOT based on vehicle signals prior to the TOR. Finally, Berghofer \textit{et al.} \cite{berghofer2019prediction} propose a regression model for TOT prediction based on driver gaze activity and driver characteristics such as age, gender, sleepiness, attitude towards highly automated driving and previous experiences with automated driving. 

Our work differs from previously proposed TOT prediction models on two counts. First, we use fine-grained descriptors of driver gaze, hand and foot activity obtained purely using non-intrusive vision sensors as inputs to our TOT prediction model. Second, we train and evaluate our models using a large \textit{real-world} dataset of take-overs captured in a conditionally autonomous vehicle. Prior work on TOT prediction has been limited to the simulator setting \cite {braunagel2017ready, lotz2018predicting, du2020predicting, hwang2020predicting, pakdamanian2021deeptake}. Berghofer \textit{et al.} \cite{berghofer2019prediction} do use a real world dataset. However, they use a 'Wizard of Oz' setting where a safety driver with access to vehicle controls plays the role of the autonomous vehicle.

\section{Datasets \& Labels}

\subsection{Controlled Data Study (CDS)}
To capture a diverse set of real-world take-overs, we conduct a large-scale study under controlled conditions. More specifically, we enlist a representative population of 89 subjects to drive a Tesla Model S testbed mounted with three driver-facing cameras that capture the gaze, hand, and foot activity of the driver. In this controlled data study (CDS), we required each subject to drive the testbed for approximately an hour in a pre-determined section of the roadway, under controlled traffic conditions. During the drive, each test subject is asked to undertake a variety of distracting secondary activities while the autopilot is engaged, following which an auditory take-over request (TOR) is issued at random intervals. This initiates the control transition during which the driver is instructed to take control of the vehicle and resume the drive. Each such transition corresponds to one take-over event, and our CDS produces 1,375 take-over events in total. 


\subsection{Annotation}
\noindent\textbf{Automated video segmentation:} Each driving session is first segmented into 30 second windows surrounding known take-over events, consisting of 20 seconds prior to the take-over request (TOR) and 10 seconds after the take-over event.  

\vspace{2mm}
\noindent\textbf{Event annotations:} 
For each 30 second clip corresponding to a take-over event, we manually annotate the three times after the take-over request corresponding to when the driver's eyes are on the road, hands are on the wheel, and foot is on the pedal. We also label the secondary activity being performed by the driver during each take-over event, assigning one of 8 possible activity labels: (1) No secondary activity, (2) talking to co-passenger, (3) eyes closed, (4) texting, (5) phone call, (6) using infotainment unit, (7) counting change, (8) reading a book or magazine. The take-over events are distributed between secondary activities as shown in Table \ref{tab:secondary}.

\begin{figure}[t]
\centering
\includegraphics[width=0.95\linewidth]{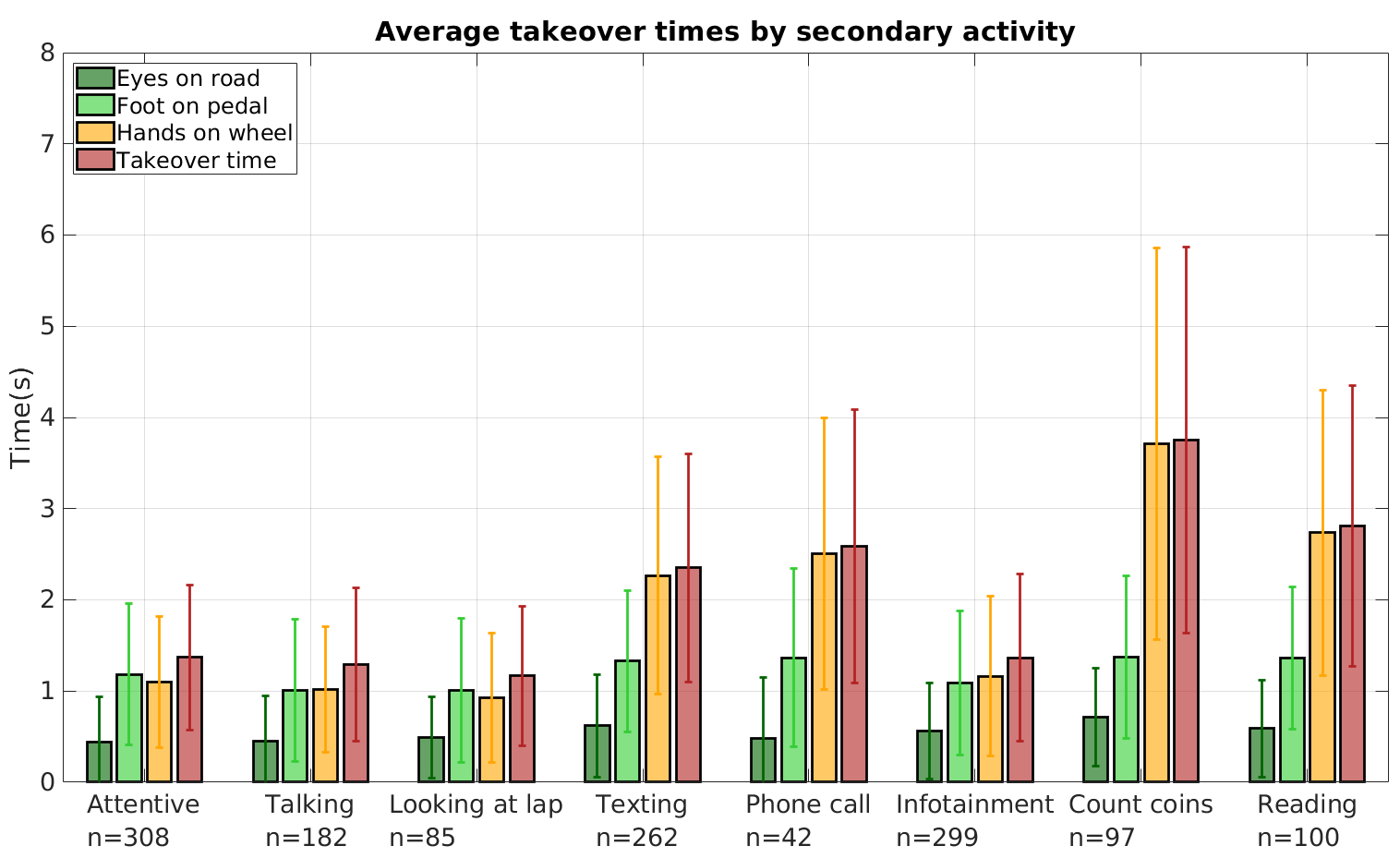}
\caption{\textbf{Take-over time statistics from the CDS:} We plot the mean values (with error bars) of the different take-over related event timings for each secondary activity.}
\label{fig:d7.1-totstats}
\end{figure}

Figure \ref{fig:d7.1-totstats} shows the average times corresponding to eyes on road, hands on wheel and foot on pedal for each of the 8 secondary activities. It also shows the overall take-over time, which is the maximum of the three markers for each event. We note that texting, phone-calls, counting change and reading correspond to longer average take-over times, as compared to talking to the co-passenger or using the infotainment unit, which can be reasonably expected. Counter to intuition, the `eyes closed behind the wheel' activity has low take-over times. This is mainly because the drivers are merely `acting' to be asleep, since actual sleep could not have been achieved given the duration and nature of each trial. We also note that the `hands on wheel' event seems to take much longer on average, as compared to eyes on road or foot on pedal. This reinforces the need for driver hand analysis, which is also a key predictor of the driver's observable readiness index (see next section). Finally, we note that for the more distracting secondary activities (reading, texting, phone calls, counting change), even the foot on pedal times are longer compared to the other secondary activities, although the secondary activities do not involve the driver's feet. Thus, there seems to be a delay corresponding to the driver shifting attention from secondary activity to the primary activity of driving.

\subsection{Data Augmentation}
\label{sec:d8-data-aug}

Takeover time data is very limited and expensive to capture and label. This is illustrated by the size of the CDS dataset (1,375 unique takeover events). This introduces challenges during training neural networks for TOT prediction, as these models typically require tens of thousands of training samples. Care must also be taken to avoid overfitting, as this is more prevalent in the limited data regime. To address these issues, we propose a new data augmentation scheme to increase the number of samples in the dataset by an order of magnitude. 

\begin{figure}[t]
\centering
\begin{subfigure}[b]{0.43\textwidth}
   \includegraphics[width=\linewidth]{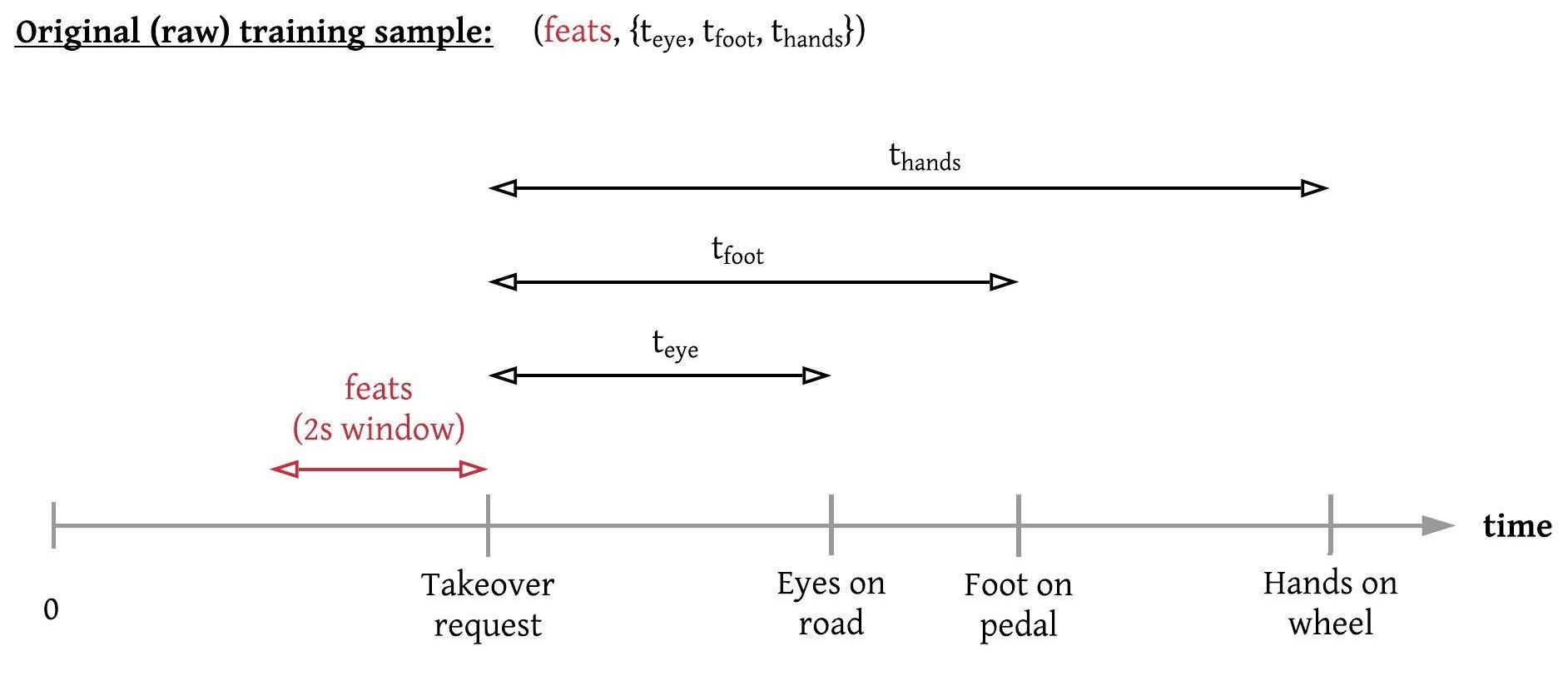}
   \caption{Raw training sample}
   \label{fig:d8-data-aug-1} 
\end{subfigure}

\vspace{5mm}

\begin{subfigure}[b]{0.48\textwidth}
   \includegraphics[width=\linewidth]{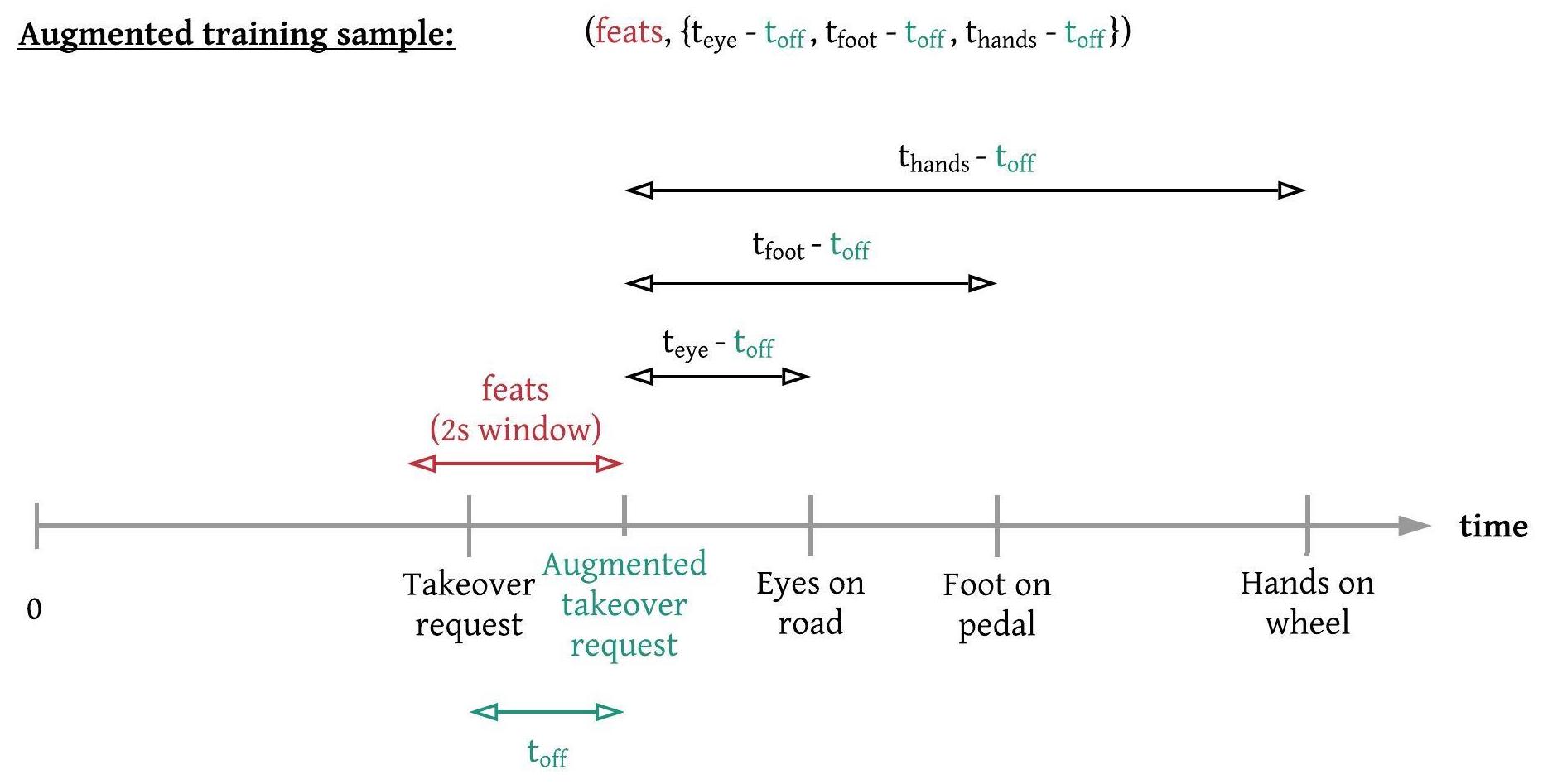}
   \caption{Augmented training sample}
   \label{fig:d8-data-aug-2}
\end{subfigure}
\caption{Illustration of TOT dataset augmentation scheme.}
\label{fig:data_aug}
\end{figure}

Figure \ref{fig:data_aug} illustrates our data augmentation scheme. We term each take-over event in the CDS dataset a \textit{raw sample}. Each raw sample has annotated timestamps corresponding to the take-over request ($t_{tor}$), as well as the time taken by the driver to get their eyes on the road ($t_{eyes}$), hands on the wheel ($t_{hands}$) and foot on the pedals ($t_{foot}$) after the TOR as shown in Figure \ref{fig:d8-data-aug-1}. We wish to learn a model that maps a 2 second window of driver activity prior to the TOR to the take-over times, $\{t_{eyes}, t_{hands}, t_{foot}\}$.

The raw samples alone are insufficient to train a machine learning model from scratch. We thus mine \textit {augmented training samples} from each takeover event as shown in Figure \ref{fig:d8-data-aug-2}. An augmented training sample is characterized by an augmented TOR at time $t_{off}$ after the actual $t_{tor}$. We use a 2 second window of driver activity before the augmented TOR as the input to the model while the corresponding takeover times are given by $\{t_{eyes} - t_{off}, t_{hands} - t_{off}, t_{foot} - t_{off}\}$. If the driver’s hands, eyes, or foot are already in position at $t_{tor}$ + $t_{off}$, the corresponding takeover time is set to 0. 

An augmented training sample maps the driver's state at an intermediate timestamp during the takeover event, to their reaction times from that timestamp. While this doesn't correspond to an actual TOR, it still serves as useful data for training our TOT prediction model as we show in Section \ref{sec:d8-ablation}. Intuitively, the driver can be expected to be less and less distracted by a non-driving activity as $t_{off}$ is increased, leading to shorter takeover times. Thus the augmented samples provide additional instances where the driver is increasingly prepared to takeover control from the vehicle. 

We capture data at a frame rate of 30 Hz. Thus, $t_{off}$ can be varied from 0 to the maximum of $\{t_{eyes}, t_{hands}, t_{foot}\}$ using increments of 1/30 seconds to yield multiple augmented samples per takeover event. The augmentation scheme is only applied to the training split. The validation and test splits are left untouched for accurate evaluation. 

\begin{table}[ht]
\centering
\begin{threeparttable}\centering
\caption{Control Transition Secondary Activity Frequency.}
\label{tab:secondary}
\begin{tabular}{@{}ccc@{}}
\hline
Secondary Activity & Number of samples & Percent\\
\hline \hline
\rowcolor[HTML]{EFEFEF}
No secondary activity & 308 & 23.0\%\\
Texting & 262 & 19.6\%\\
\rowcolor[HTML]{EFEFEF}
Infotainment unit & 262 & 19.6\%\\
Talking to passenger & 182 & 13.6\%\\
\rowcolor[HTML]{EFEFEF}
Reading book or magazine & 100 & 7.5\% \\
Counting coins & 97 & 7.2\%\\
\rowcolor[HTML]{EFEFEF}
Eyes closed/Looking at lap & 85 & 6.4\%\\
Phone call & 42 & 3.1\%\\
\bottomrule
\end{tabular}
\end{threeparttable}
\end{table}

\section{Models \& Algorithms for Predicting Takeover Times}

\begin{figure}[t]
  \centering
  \includegraphics[width=.95\linewidth]{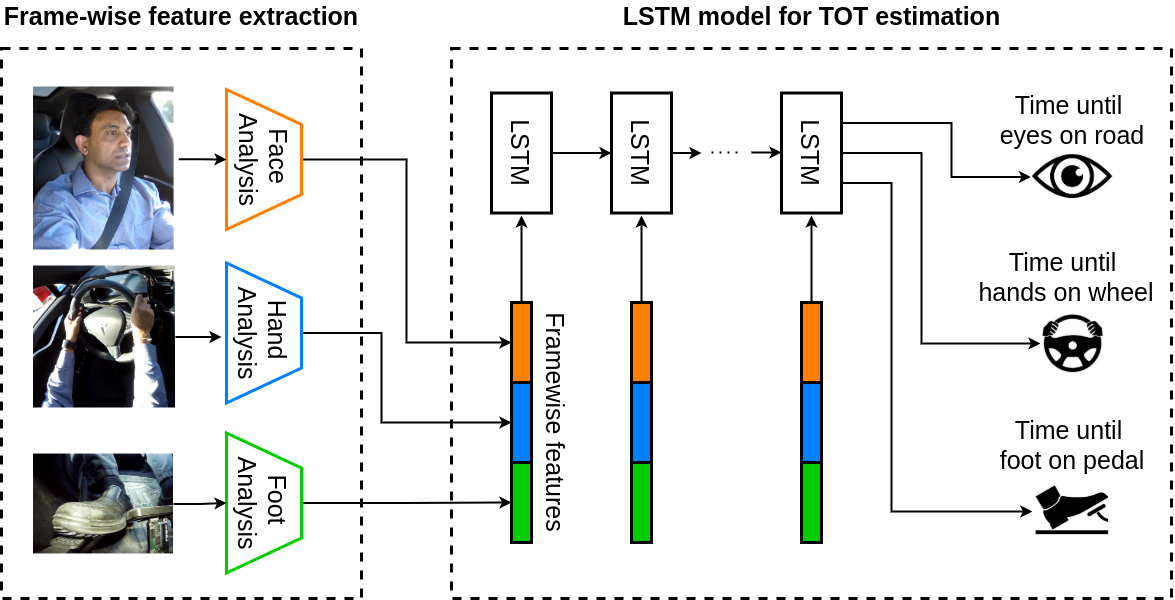}
  \caption{\textbf{Overview of the proposed approach:} We extract frame-wise descriptors of driver gaze, hand and foot activity. We propose an LSTM model for predicting TOT based on a sequence of the extracted features over a 2 second window.}
  \label{fig:d5-framework_tot}
\end{figure}

It is important to preserve both the diverse and sequential nature of all features related to driver behavior while designing a holistic take-over time (TOT) prediction framework. High level tasks such as TOT prediction are influenced by low level driver behaviors, both in the short and medium to long term. Figure \ref{fig:d5-framework_tot} provides an overview of our proposed approach for estimating TOT. Our approach consists of two major components. The first component is a set of convolutional neural networks (CNNs) for extracting frame-wise descriptors of driver gaze, hand and foot activity from the raw camera feed. We describe these is greater detail in section \ref{sec:d5-tot-framewise}. The second component is an LSTM model for estimating TOT based on a sequence of frame-wise features over a pre-defined time window. We describe the different variants of our LSTM based models in section \ref{sec:d8-tot-models}. 


\subsection{Frame-wise feature extraction}\label{sec:d5-tot-framewise}
\vspace{1mm}
\noindent\textbf{Gaze activity:} We use the model proposed by Vora \textit{et al}. \cite{vora2017generalizing} for driver gaze analysis. The inputs to the model are frames from the face camera. We use a face detector \cite{yuen2016looking} for localizing the driver's eyes. A cropped bounding box around the driver's eyes is passed through a CNN, which outputs the driver's gaze zone. We consider 8 gaze zones: \{forward, left mirror, lap, speedometer, infotainment unit, rear-view mirror, right mirror, over the shoulder\}. The CNN outputs frame-wise probabilities for each gaze zone. We use this 8 dimensional vector to represent driver's gaze features.     

\vspace{1mm}
\noindent\textbf{Hand activity:} We use the model proposed by Yuen and Trivedi \cite{yuen2018looking} for driver hand analysis. The model localizes the elbow and wrist joints of the driver using part affinity fields \cite{cao2017realtime}. A cropped bounding box around the driver's wrist is passed through a CNN to output probabilities corresponding to 6 hand activities for each hand:  \{on lap, in air, hovering over steering wheel, on steering wheel, on cupholder, interacting with infotainment unit\}. We extend the model to additionally output hand-held object probabilities. We consider 7 object categories: \{no-object, phone, tablet, food, beverage, book, other\}. 
By running the models on images from a stereo camera pair, we also obtain 3-d coordinates for the driver's wrist locations and the steering wheel using triangulation. We then calculate the distance of each hand (wrist) of the driver to the steering wheel in 3-d.
The hand activity probabilities, hand object probabilities and 3-d distance to steering wheel together form the hand activity features for each frame. 

\vspace{1mm}
\noindent\textbf{Foot activity:} We use the model proposed by Rangesh and Trivedi \cite{rangesh2019forced} for driver foot analysis. Each frame from the foot camera feed is passed through a CNN to output probabilities over 5 foot activity classes: \{away from pedal, on brake, on gas, hovering over brake, hovering over gas\}. These probabilities represent the foot activity features for each frame.  

\subsection{LSTM models for take-over time prediction}
\label{sec:d8-tot-models}


\vspace{2mm}
\noindent\textbf{Baseline LSTM:} This is the simplest (baseline) version of all TOT models.
The input features are first transformed using a fully-connected (FC) layer of size 16 (plus non-linearity), which is then fed to an LSTM with a hidden state of size 32 at each timestep. 
The LSTM layer receives the transformed input features at each timestep and updates its internal representation known as the hidden state. In all our experiments, we choose a 2 second window of features as input to our models.
After 2 seconds worth of inputs and updates, the hidden state of the LSTM after the latest timestep is passed through an output transformation (FC layer plus non-linearity) to predict the three times of interest.

We apply a simple $L1$ loss to train this network. Let $o_e$, $o_f$, and $o_h$ be the outputs produced by the model. Assuming $t_e$, $t_f$, and $t_h$ are the target eyes on road time, foot on pedal time, and hands on wheel time respectively, the total loss is:

\begin{equation}
\label{eq:l1_loss}
    \mathcal{L} = \frac{1}{N} \sum_{i=1}^{N} |t_e^i - o_e^i| + \frac{1}{N} \sum_{i=1}^{N} |t_f^i - o_f^i| + \frac{1}{N} \sum_{i=1}^{N} |t_h^i - o_h^i|.
\end{equation}

The entire model is trained using an Adam optimizer with a learning rate of $0.001$ for $10$ epochs.

\begin{figure}[t]
\centering
\includegraphics[width=0.9\linewidth]{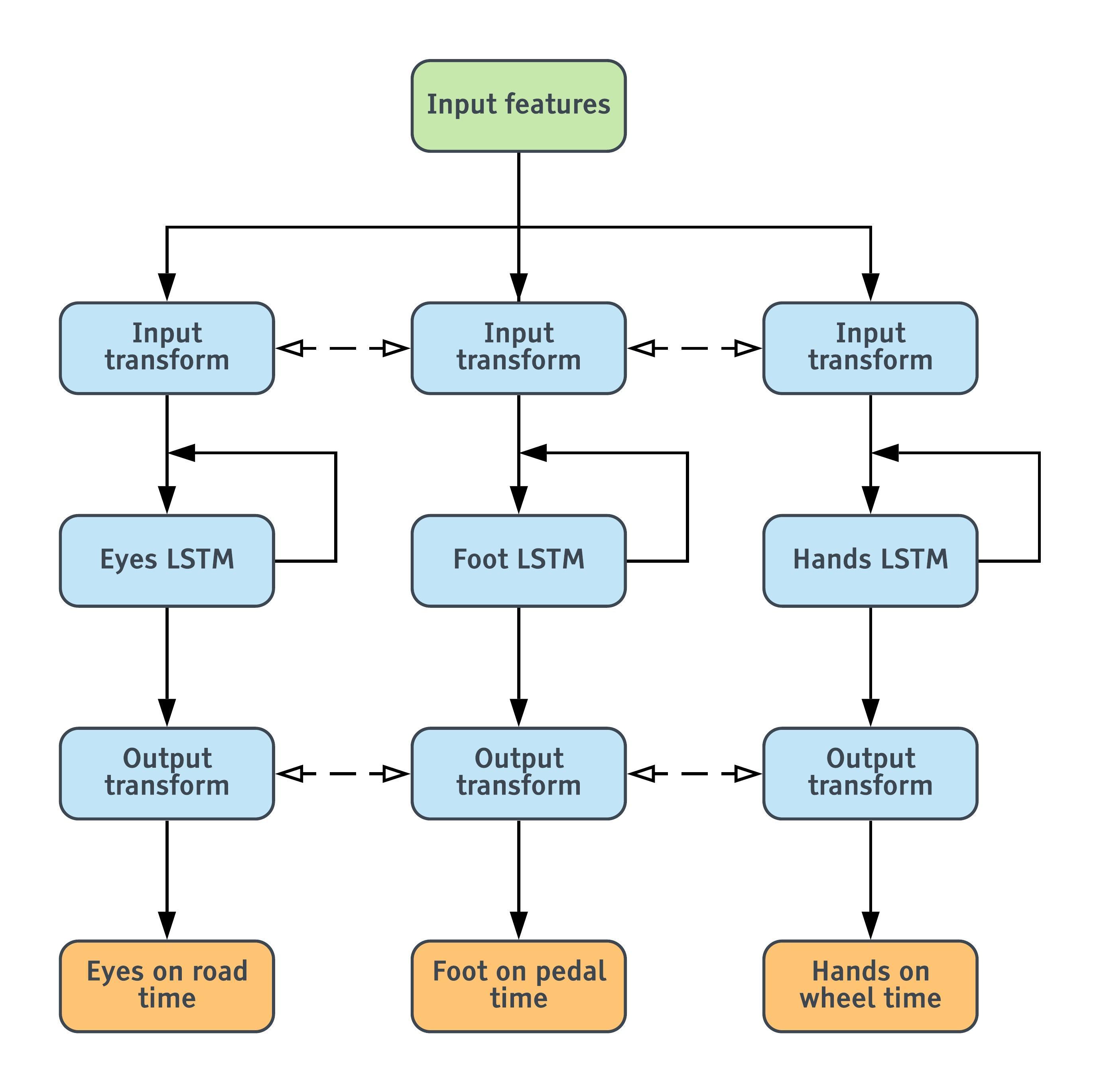}
\caption{Independent LSTMs model architecture.}
\label{fig:d8-id-lstms}
\end{figure}

\vspace{2mm}
\noindent\textbf{Independent LSTMs:} Figure \ref{fig:d8-id-lstms} shows the independent LSTM model architecture. This model is the same as the baseline LSTM model, except for one major difference: each target output time has its own independent LSTM.
The reasoning behind this is to accommodate different hidden state update rates for different driver behaviors, for example – eyes on road behavior is generally faster (short term) than hands on wheel behavior (mid/long term). Having multiple independent LSTMs allows each one to update at different rates, thereby capturing short/mid/long term behaviours separately.

Although each branch has its own LSTM cell, the input and output transformations are still shared between the three LSTMs as the feature inputs to the three branches are the same. This tends to reduce overfitting based on our experiments.

We use the identical loss (eq. \ref{eq:l1_loss}) and optimizer settings as the baseline LSTM for training the independent LSTMs model. 





\vspace{2mm}
\noindent\textbf{LSTM with Multi-modal Outputs:} This model is largely based on the baseline LSTM with one addition: \textit{multi-modal outputs}.
Instead of just producing one output for each of the three targets, we output $K (=3)$ outputs per target and their associated probabilities. We do this to model the inherent multi-modality and subjectiveness of takeover times. For example, given similar history of behavior, one driver may respond faster in taking control of the vehicle than another. Producing multiple probable outputs (and their probabilities) could possibly address this ambiguity and provide more usable information to any downstream controller.

Unlike the previous models, this model is trained using a minimum of $K$ loss, where $L1$ losses are only applied to the output modes closest to the ground truth target. Additionally, the output probabilities are refined using cross-entropy. Let $o_e(k)$, $o_f(k)$, $o_h(k)$ and $q(k)$ denote the $k^{th}$ set of outputs and corresponding probability produced by the model. Assuming $t_e$, $t_f$, and $t_h$ are the target eyes on road time, foot on pedal time, and hands on wheel time respectively, the total loss is:

\begin{multline}
\label{eq:mm_loss}
    \mathcal{L} = \frac{1}{N} \sum_{i=1}^{N} \min_k \big(|t_e^i - o_e^i(k)| + |t_f^i - o_f^i(k)| + |t_h^i - o_h^i(k)|\big)\\ 
    - \lambda \frac{1}{N} \sum_{i=1}^{N} \sum_{k=1}^{K} p^i(k) \log(q^i(k)),
\end{multline}
where $p^i(k)$ is a one-hot categorical probability distribution given by the indicator function,
\begin{equation}
\resizebox{0.91\hsize}{!}{%
    $p^i(k) = \mathbbm{1}\bigg(\argmin_l \big(|t_e^i - o_e^i(l)| + |t_f^i - o_f^i(l)| + |t_h^i - o_h^i(l)|\big) = k\bigg),$
    }
\end{equation}
and $\lambda$ is a coefficient used for relatively weighting the L1 and cross-entropy losses.

As before, the entire model is trained using an Adam optimizer with a learning rate of $0.001$ for $10$ epochs. We use $\lambda=1$ for simplicity. 

\begin{figure}[t]
\centering
\includegraphics[width=0.9\linewidth]{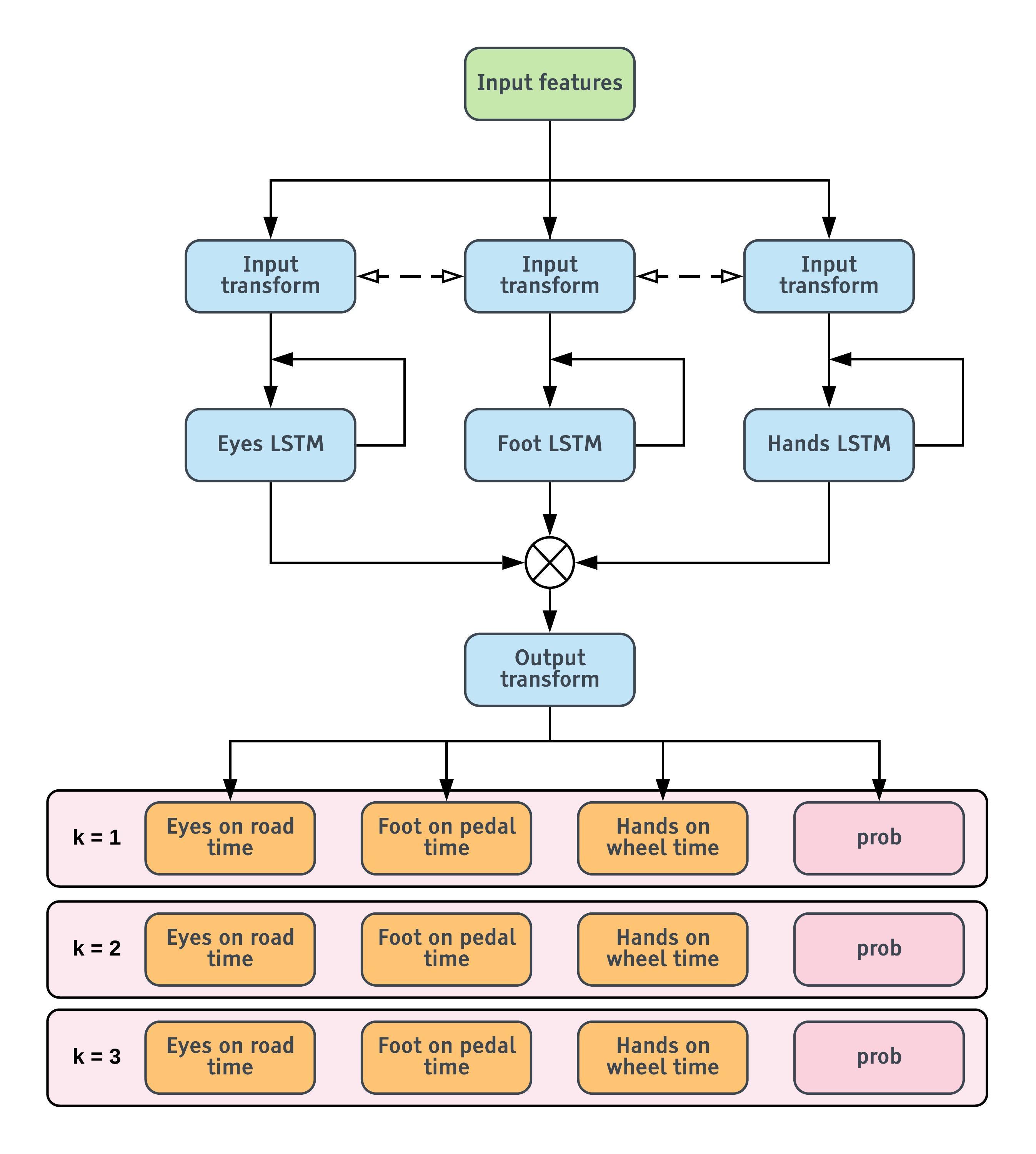}
\caption{Independent LSTMs with multi-modal outputs model architecture.}
\label{fig:d8-id-lstms-mm}
\end{figure}

\vspace{2mm}
\noindent\textbf{Independent LSTMs with Multi-modal Outputs:} The final proposed model uses a combination of independent LSTMs and multi-modal outputs described before.
One difference to the original independent LSTMs model is that we now concatenate the hidden states of all three LSTMs and transform them together to produce the target outputs.
This is done because probabilities are assigned to the joint of all three target times, and thus need to be operated on together.



We use the identical loss (equation \ref{eq:mm_loss}) and optimizer settings as the LSTM with multi-modal outputs for training the independent LSTMs with multi-modal outputs.

\section{Experiments \& Evaluation}
\label{sec:d8-ablation}


\begin{table*}[ht]
\centering
\begin{threeparttable}\centering
\caption{Estimation errors for different times of interest on the CDS validation set.} 
\label{tab:model-comp}
\begin{tabular}{@{}cccccc@{}}
\hline
\begin{tabular}[c]{@{}c@{}}Model\\ type (s)\end{tabular} & 
\begin{tabular}[c]{@{}c@{}}Overall\\ MAE (s)\end{tabular} & 
\begin{tabular}[c]{@{}c@{}}Eyes on road\\ MAE (s)\end{tabular} & 
\begin{tabular}[c]{@{}c@{}}Foot on pedal\\ MAE (s)\end{tabular} & 
\begin{tabular}[c]{@{}c@{}}Hands on wheel\\ MAE (s)\end{tabular} & 
\begin{tabular}[c]{@{}c@{}}Takeover time\\ MAE (s)\end{tabular}\\
\hline \hline
\rowcolor[HTML]{EFEFEF}
Constant prediction (Max over train set stats)             & 3.9271 & 2.4540 & 2.9880 & 6.3392 & 6.1969 \\
LSTM\tnote{1}             & 0.5104 & 0.3353 & 0.5029 & 0.7126 & 0.8098 \\
\rowcolor[HTML]{EFEFEF}ID LSTMs\tnote{2}         & \textbf{0.5073} & \textbf{0.3266} & \textbf{0.4841} & \textbf{0.7113} & \textbf{0.7912} \\
LSTM + MM\tnote{3}        & 0.5589 & 0.3582 & 0.5262 & 0.7921 & 0.8908 \\
\rowcolor[HTML]{EFEFEF}ID LSTMs + MM             & 0.5319 & 0.3415 & 0.5019 & 0.7524 & 0.8441 \\
\hline
LSTM + MM (best of K)     & 0.3921 & 0.2393 & 0.4204 & 0.5167 & 0.6265 \\
\rowcolor[HTML]{EFEFEF}ID LSTMs + MM (best of K) & 0.3911 & 0.2344 & 0.3875 & 0.5513 & 0.6586 \\
\bottomrule
\end{tabular}
\begin{tablenotes}
    \item[1] baseline LSTM model
    \item[2] Independent LSTMs
    \item[3] Multi-modal outputs (with $K = 3$ modes)
 \end{tablenotes}
\end{threeparttable}
\end{table*}

\vspace{2mm}
\noindent\textbf{Comparison of LSTM models for TOT prediction:}
First, we conduct an experiment to assess the effects of different model architectures. All proposed models (from Section~\ref{sec:d8-tot-models}) were trained on CDS train set with augmented data, and then evaluated on the validation set. We use individual and overall mean absolute errors (MAEs) as metrics for comparison. Table~\ref{tab:model-comp} contains results from this experiment. In addition to the LSTM models, as a sanity check, we include a simple baseline that always predicts a constant value for all take-over time markers, corresponding to the maximum value for each marker from the train set. 

From these results, we note that all LSTM models considerably outperform the constant value baseline showing that there is a learnable signal in the data and the usefulness of using a machine learning model.
We observe that the independent LSTMs model consistently outperforms other models. At first glance, the multi-modal models tend to perform worse than the ones without multi-modal outputs. To further analyze the source of these errors, we provide the \textit{best-of-K} MAEs for these models in Table~\ref{tab:model-comp}. The \textit{best-of-K} MAEs simply mean that instead of choosing the most probable set of predictions for error calculation, we use the set that produces the least error i.e. assume perfect classification. The \textit{best-of-K} numbers are vastly superior to the ones without multi-modal outputs. This indicates that in most cases, at least one of $K (=3)$ sets of predictions is highly accurate. However, accurate probability assignment for these $K$ modes (i.e. classification) remains error-prone. Nevertheless, we believe that having multiple probable outputs instead of one less accurate one could be beneficial for downstream controllers.

\begin{table*}[ht]
\centering
\begin{threeparttable}\centering
\caption{Estimation errors for different times of interest on the CDS validation set when trained on a variety of datasets.}
\label{tab:data-comp}
\begin{tabular}{@{}cccccc@{}}
\hline
\begin{tabular}[c]{@{}c@{}}Training\\ dataset (s)\end{tabular} & 
\begin{tabular}[c]{@{}c@{}}Overall\\ MAE (s)\end{tabular} & 
\begin{tabular}[c]{@{}c@{}}Eyes on road\\ MAE (s)\end{tabular} & 
\begin{tabular}[c]{@{}c@{}}Foot on pedal\\ MAE (s)\end{tabular} & 
\begin{tabular}[c]{@{}c@{}}Hands on wheel\\ MAE (s)\end{tabular} & 
\begin{tabular}[c]{@{}c@{}}Takeover time\\ MAE (s)\end{tabular}\\
\hline \hline
\rowcolor[HTML]{EFEFEF}
CDS (R)\tnote{1}                            & 0.5799 & 0.3676 & 0.5435 & 0.8285 & 0.8576 \\
CDS (A)\tnote{2}                            & \textbf{0.5073} & 0.3266 & 0.4841 & 0.7113 & 0.7912 \\
\rowcolor[HTML]{EFEFEF}
ORI\tnote{3} $\rightarrow$ CDS (A) (Fig.~\ref{fig:d8-transfer}) & 0.5184 & \textbf{0.3246} & 0.5182 & \textbf{0.7054} & \textbf{0.7729} \\
\bottomrule
\end{tabular}
\begin{tablenotes}
    \item[1] raw dataset
    \item[2] augmented dataset
    \item[3] ORI estimation dataset
 \end{tablenotes}
\end{threeparttable}
\end{table*}

\vspace{2mm}
\noindent\textbf{Effect of data augmentation and transfer learning:}
Next, we conduct experiments to assess the effects of our data augmentation and transfer learning schemes. To isolate these effects, we use the same ID LSTMs model for all experiments. We compare the following training schemes:

\begin{itemize}
    \item \textbf{CDS (R):} First, as a baseline, we train a model purely using the raw CDS data without augmentation.
    \item \textbf{CDS (A):} Next, we train a model using the augmented training dataset using the augmentation scheme described in section \ref{sec:d8-data-aug}. The raw dataset contained 1,375 samples, which we augment to 47,461 datapoints. 
    \item \textbf{ORI $\rightarrow$ CDS (A):} Finally, we consider a model pre-trained to estimate the observable take-over readiness index (ORI) proposed in \cite{deo2019looking}. The ground truth ORI values are obtained via subjective ratings assigned by multiple human observers rating how ready a driver is to take-over control from the vehicle based on the past two seconds of video feed from the driver facing cameras. The ratings are normalized and averaged to account for rater bias as described in \cite{deo2019looking}.
\end{itemize}

Results from these experiments are presented in Table~\ref{tab:data-comp}. As before, we use individual and overall mean absolute errors (MAEs) as metrics for comparison. 

From Table~\ref{tab:data-comp}, we notice that training on the augmented dataset (as proposed in Section~\ref{sec:d8-data-aug}) consistently and considerably improves performance as compared to the raw dataset. We believe that doing so prevents overfitting, provides regularization, smooths the outputs of model, and adds new training samples that would be cumbersome or impossible to capture. 



Finally, we observe that training the model for observable readiness index (ORI) estimation \cite{deo2019looking}, followed by transfer learning on TOT prediction improves some metrics. This highlights the commonality between the two tasks - features from learning one task can improve performance in the other. 


\begin{table*}[ht]
\centering
\begin{threeparttable}\centering
\caption{Estimation errors for different times of interest on the CDS validation set for a variety of feature combinations.}
\label{tab:feat-comp}
\begin{tabular}{@{}cccccccccc@{}}
\toprule
\multicolumn{5}{c}{Features} & \\ \cmidrule(r){1-5}
Foot&
Gaze&
Hands Activities&
Hands Distances&
Held Objects&
\multirow{-2}{*}{\begin{tabular}[c]{@{}c@{}}Overall\\ MAE (s)\end{tabular}} & \multirow{-2}{*}{\begin{tabular}[c]{@{}c@{}}Eyes on road\\ MAE (s)\end{tabular}} & \multirow{-2}{*}{\begin{tabular}[c]{@{}c@{}}Foot on pedal\\ MAE (s)\end{tabular}} & \multirow{-2}{*}{\begin{tabular}[c]{@{}c@{}}Hands on wheel\\ MAE (s)\end{tabular}} & \multirow{-2}{*}{\begin{tabular}[c]{@{}c@{}}Takeover time\\ MAE (s)\end{tabular}}\\
\hline \hline
\rowcolor[HTML]{EFEFEF}
 \ding{51} & & & & & 0.5735 & 0.3587 & 0.5018 & 0.8599 & 0.8856 \\
 & \ding{51} & & & & 0.5811 & 0.3332 & 0.5690 & 0.8411 & 0.8837 \\
\rowcolor[HTML]{EFEFEF}
 & & \ding{51} & & & 0.5560 & 0.3729 & 0.5384 & 0.7565 & 0.9012 \\
 & & \ding{51} & \ding{51} & & 0.5420 & 0.3783 & 0.5109 & 0.7369 & 0.8315 \\
\rowcolor[HTML]{EFEFEF}
 & & \ding{51} & & \ding{51} & 0.5217 & 0.3702 & 0.4973 & 0.7177 & 0.8621 \\
 & & \ding{51} & \ding{51} & \ding{51} & 0.5182 & 0.3747 & 0.4857 & 0.7141 & 0.7983 \\
\rowcolor[HTML]{EFEFEF}
 & \ding{51} & \ding{51} & & \ding{51} & 0.5202 & 0.3244 & 0.5220 & 0.7163 & 0.7920 \\
 & \ding{51} & \ding{51} & \ding{51} & \ding{51} & 0.5213 & 0.3299 & 0.5124 & 0.7215 & 0.7921 \\
 \rowcolor[HTML]{EFEFEF}
\ding{51} & \ding{51} & \ding{51} & \ding{51} & & 0.5384 & \textbf{0.3222} & 0.5059 & 0.7870 & 0.8475 \\
\ding{51} & \ding{51} & \ding{51} & & \ding{51} & 0.5088 & 0.3277 & 0.5074 & 0.7144 & 0.7918 \\
\rowcolor[HTML]{EFEFEF}
\ding{51} & \ding{51} & \ding{51} & \ding{51} & \ding{51} & \textbf{0.5073} & 0.3266 & \textbf{0.4841} & \textbf{0.7113} & \textbf{0.7912} \\
\bottomrule
\end{tabular}
\end{threeparttable}
\end{table*}

\vspace{2mm}
\noindent \textbf{Effect of hand, gaze and foot activity features:} 
Finally, we conduct an experiment to assess the relative importance of different input features and their combinations. To isolate effects from features, we train the same ID LSTMs model with different input feature combinations. We use individual and overall mean absolute errors (MAEs) as metrics for comparison. Table~\ref{tab:feat-comp} contains results from this experiment.

We notice that hand features are the most important, followed by foot and gaze features respectively. This might be because gaze dynamics are relatively predictable during takeovers as the first thing drivers tend to do is look at the road to assess the situation, leading to less variance in eyes-on-road behavior. Next, we notice that adding more informative hand feature like 3D distances to the steering wheel and hand-object information improves the performance further. Hand-objects in particular seem to vastly improve the performance in general. This makes sense as hand-objects are the strongest cue related the secondary activities of drivers. Adding stereo hand features improves the results, but not by much. Adding foot features also tends to reduce the errors considerably, illustrating the importance of having a foot camera.

In conclusion, one could get close to peak performance by utilizing 3 cameras - 1 foot, 1 hand, and 1 face camera respectively. Hand features are most informative, followed by foot and  gaze features respectively. 




\begin{table*}[t]
\centering
\begin{threeparttable}\centering
\caption{Estimation errors for different models on the takeover time test set.}
\label{tab:test-mae}
\begin{tabular}{@{}cccccc@{}}
\hline
\begin{tabular}[c]{@{}c@{}}Model\\ type (s)\end{tabular} & 
\begin{tabular}[c]{@{}c@{}}Overall\\ MAE (s)\end{tabular} & 
\begin{tabular}[c]{@{}c@{}}Eyes on road\\ MAE (s)\end{tabular} & 
\begin{tabular}[c]{@{}c@{}}Foot on pedal\\ MAE (s)\end{tabular} & 
\begin{tabular}[c]{@{}c@{}}Hands on wheel\\ MAE (s)\end{tabular} & 
\begin{tabular}[c]{@{}c@{}}Takeover time\\ MAE (s)\end{tabular}\\
\hline \hline
\rowcolor[HTML]{EFEFEF}
Constant prediction (Max over train set stats)             & 4.0835  & 2.6790  & 3.1540 & 6.4175 & 6.2073 \\

LSTM\tnote{1}              & 0.5242 & \textbf{0.2365} & 0.5007 & 0.8710 & 0.9457 \\
\rowcolor[HTML]{EFEFEF}
ID LSTMs\tnote{2}          & \textbf{0.5208} & 0.2497 & \textbf{0.4650} & \textbf{0.8055} & \textbf{0.9144} \\
LSTM + MM\tnote{3}         & 0.5339 & 0.2635 & 0.5265 & 0.8117 & 0.9307 \\
\rowcolor[HTML]{EFEFEF}
ID LSTMs + MM              & 0.5526 & 0.2665 & 0.5180 & 0.8734 & 0.9418 \\
\hline
ID LSTMs (75\%\tnote{4}\ ) & 0.5348 & 0.2557 & 0.5013 & 0.8474 & 0.9779 \\
\rowcolor[HTML]{EFEFEF}
ID LSTMs (90\%\tnote{5}\ ) & 0.5282 & 0.2514 & 0.4851 & 0.8482 & 0.9424 \\
\bottomrule
\end{tabular}
\begin{tablenotes}
    \item[1] baseline LSTM model
    \item[2] Independent LSTMs
    \item[3] Multi-modal outputs (with $K = 3$ modes)
    \item[4] 75\% of the dataset used for training
    \item[5] 90\% of the dataset used for training
 \end{tablenotes}
\end{threeparttable}
\end{table*}

\label{sec:d8-results}
\vspace{2mm}
\noindent\textbf{Quantitative results on test set:} In this section, we present quantitative error metrics on the held out test set, separate from the validation set, for all proposed models in Table~\ref{tab:test-mae}. As before, we see that ID LSTMs is the best performing model. We also notice that hands-on-wheel MAEs are usually the largest owing to large variance in hand behaviors, and large absolute values associated with hands-on-wheel time.

We also show results for ID LSTMs when trained on 75\% and 90\% of available training data. This helps us gauge the expected improvement in performance as more training data is added. Based on the numbers presented in Table~\ref{tab:test-mae}, we can expect meager improvements as more data is added. This indicates a case of diminishing returns.

\section{Concluding Remarks}
This paper presented one of the largest real-world studies on takeover time prediction and control transitions in general. We introduced a dataset of take-over events captured via controlled driving studies in a commercially available partially autonomous vehicle, with a large pool of test subjects performing a variety of secondary activities prior to the control transition. We proposed a machine learning model for take-over time prediction based on driver gaze, hand and foot activity prior to the issue of take-over requests. We also proposed a data augmentation and transfer learning scheme for best utilizing the limited number of take-over events in our dataset. Our experiments show that our model can reliably predict takeover times for various secondary activities being performed by the drivers. In particular, we showed the usefulness of analyzing driver hand, foot and gaze activity prior to issuing the take-over request. We also showed the utility of our transfer learning and data augmentation schemes for best utilizing limited training data with control transitions. We believe that this study outlines the sensors, datasets, methods and models that can benefit the intermediate stages of automation by accurately assessing driver behavior, and predicting takeover times - both of which can be used to smoothly transfer control between human and automation.



\section{Acknowledgments}
We would like to thank Toyota Collaborative Safety Research Center (CSRC) and our other industry sponsors for their generous and continued support. We would also like to thank our colleagues at Laboratory for Intelligent and Safe Automobiles (LISA), UC San Diego and National Advanced Driving Simulator (NADS) University of Iowa, for their useful inputs and help in collecting and labeling the dataset. 

\bibliographystyle{IEEEtran}

\bibliography{references}

\begin{thebibliography}{10}
\providecommand{\url}[1]{#1}
\csname url@rmstyle\endcsname
\providecommand{\newblock}{\relax}
\providecommand{\bibinfo}[2]{#2}
\providecommand\BIBentrySTDinterwordspacing{\spaceskip=0pt\relax}
\providecommand\BIBentryALTinterwordstretchfactor{4}
\providecommand\BIBentryALTinterwordspacing{\spaceskip=\fontdimen2\font plus
\BIBentryALTinterwordstretchfactor\fontdimen3\font minus
  \fontdimen4\font\relax}
\providecommand\BIBforeignlanguage[2]{{%
\expandafter\ifx\csname l@#1\endcsname\relax
\typeout{** WARNING: IEEEtran.bst: No hyphenation pattern has been}%
\typeout{** loaded for the language `#1'. Using the pattern for}%
\typeout{** the default language instead.}%
\else
\language=\csname l@#1\endcsname
\fi
#2}}

\bibitem{kircher2017minimum}
K.~Kircher and C.~Ahlstrom, ``Minimum required attention: a human-centered
  approach to driver inattention,'' \emph{Human factors}, vol.~59, no.~3, pp.
  471--484, 2017.

\bibitem{jensen2010studying}
B.~S. Jensen, M.~B. Skov, and N.~Thiruravichandran, ``Studying driver attention
  and behaviour for three configurations of gps navigation in real traffic
  driving,'' in \emph{Proceedings of the SIGCHI Conference on Human Factors in
  Computing Systems}, 2010, pp. 1271--1280.

\bibitem{tice2021driver}
P.~Tice, S.~Dey~Tirtha, and N.~Eluru, ``Driver attention and the built
  environment initial, findings from a naturalistic driving study,'' in
  \emph{Proceedings of the Human Factors and Ergonomics Society Annual
  Meeting}, vol.~65, no.~1.\hskip 1em plus 0.5em minus 0.4em\relax SAGE
  Publications Sage CA: Los Angeles, CA, 2021, pp. 1077--1081.

\bibitem{gaspar2019effect}
J.~Gaspar and C.~Carney, ``The effect of partial automation on driver
  attention: A naturalistic driving study,'' \emph{Human factors}, vol.~61,
  no.~8, pp. 1261--1276, 2019.

\bibitem{benedetto2011driver}
S.~Benedetto, M.~Pedrotti, L.~Minin, T.~Baccino, A.~Re, and R.~Montanari,
  ``Driver workload and eye blink duration,'' \emph{Transportation research
  part F: traffic psychology and behaviour}, vol.~14, no.~3, pp. 199--208,
  2011.

\bibitem{ojstervsek2019eye}
T.~C. Ojster{\v{s}}ek, ``Eye tracking use in researching driver distraction: A
  scientometric and qualitative literature review approach,'' \emph{Journal of
  Eye Movement Research}, vol.~12, no.~3, 2019.

\bibitem{isaza2019dynamic}
C.~Isaza, K.~Anaya, C.~Fuentes-Silva, J.~P.~Z. de~Paz, A.~Rizzo, and A.-I.
  Garcia-Moreno, ``Dynamic set point model for driver alert state using digital
  image processing,'' \emph{Multimedia Tools and Applications}, vol.~78,
  no.~14, pp. 19\,543--19\,563, 2019.

\bibitem{eastwood2012unengaged}
J.~D. Eastwood, A.~Frischen, M.~J. Fenske, and D.~Smilek, ``The unengaged mind:
  Defining boredom in terms of attention,'' \emph{Perspectives on Psychological
  Science}, vol.~7, no.~5, pp. 482--495, 2012.

\bibitem{charlton2011driving}
S.~G. Charlton and N.~J. Starkey, ``Driving without awareness: The effects of
  practice and automaticity on attention and driving,'' \emph{Transportation
  research part F: traffic psychology and behaviour}, vol.~14, no.~6, pp.
  456--471, 2011.

\bibitem{bernstein2015texting}
J.~J. Bernstein and J.~Bernstein, ``Texting at the light and other forms of
  device distraction behind the wheel,'' \emph{BMC public health}, vol.~15,
  no.~1, pp. 1--5, 2015.

\bibitem{nhtsa}
\BIBentryALTinterwordspacing
``Automated vehicles for safety.'' [Online]. Available:
  \url{https://www.nhtsa.gov/technology-innovation/automated-vehicles-safety}
\BIBentrySTDinterwordspacing

\bibitem{williams2021automated}
B.~Williams, ``Automated driving levels,'' 2021.

\bibitem{casner2016challenges}
S.~M. Casner, E.~L. Hutchins, and D.~Norman, ``The challenges of partially
  automated driving,'' \emph{Communications of the ACM}, vol.~59, no.~5, pp.
  70--77, 2016.

\bibitem{kyriakidis2017human}
M.~Kyriakidis, J.~C. de~Winter, N.~Stanton, T.~Bellet, B.~van Arem,
  K.~Brookhuis, M.~H. Martens, K.~Bengler, J.~Andersson, N.~Merat,
  \emph{et~al.}, ``A human factors perspective on automated driving,''
  \emph{Theoretical Issues in Ergonomics Science}, pp. 1--27, 2017.

\bibitem{warm2008vigilance}
J.~S. Warm, R.~Parasuraman, and G.~Matthews, ``Vigilance requires hard mental
  work and is stressful,'' \emph{Human factors}, vol.~50, no.~3, pp. 433--441,
  2008.

\bibitem{rangesh2021autonomous}
A.~Rangesh, N.~Deo, R.~Greer, P.~Gunaratne, and M.~M. Trivedi, ``Autonomous
  vehicles that alert humans to take-over controls: Modeling with real-world
  data,'' \emph{arXiv preprint arXiv:2104.11489}, 2021.

\bibitem{deo2019looking}
N.~Deo and M.~M. Trivedi, ``Looking at the driver/rider in autonomous vehicles
  to predict take-over readiness,'' \emph{IEEE Transactions on Intelligent
  Vehicles}, vol.~5, no.~1, pp. 41--52, 2019.

\bibitem{murphy2008hyhope}
E.~Murphy-Chutorian and M.~M. Trivedi, ``Hyhope: Hybrid head orientation and
  position estimation for vision-based driver head tracking,'' in \emph{2008
  IEEE Intelligent Vehicles Symposium}.\hskip 1em plus 0.5em minus 0.4em\relax
  IEEE, 2008, pp. 512--517.

\bibitem{tawari2014robust}
A.~Tawari and M.~M. Trivedi, ``Robust and continuous estimation of driver gaze
  zone by dynamic analysis of multiple face videos,'' in \emph{Intelligent
  Vehicles Symposium Proceedings, 2014 IEEE}.\hskip 1em plus 0.5em minus
  0.4em\relax IEEE, 2014, pp. 344--349.

\bibitem{tawari2014driver}
A.~Tawari, K.~H. Chen, and M.~M. Trivedi, ``Where is the driver looking:
  Analysis of head, eye and iris for robust gaze zone estimation,'' in
  \emph{Intelligent Transportation Systems (ITSC), 2014 IEEE 17th International
  Conference on}.\hskip 1em plus 0.5em minus 0.4em\relax IEEE, 2014, pp.
  988--994.

\bibitem{lee2011real}
S.~J. Lee, J.~Jo, H.~G. Jung, K.~R. Park, and J.~Kim, ``Real-time gaze
  estimator based on driver's head orientation for forward collision warning
  system,'' \emph{IEEE Transactions on Intelligent Transportation Systems},
  vol.~12, no.~1, pp. 254--267, 2011.

\bibitem{vasli2016driver}
B.~Vasli, S.~Martin, and M.~M. Trivedi, ``On driver gaze estimation:
  Explorations and fusion of geometric and data driven approaches,'' in
  \emph{Intelligent Transportation Systems (ITSC), 2016 IEEE 19th International
  Conference on}.\hskip 1em plus 0.5em minus 0.4em\relax IEEE, 2016, pp.
  655--660.

\bibitem{fridman2016driver}
L.~Fridman, P.~Langhans, J.~Lee, and B.~Reimer, ``Driver gaze region estimation
  without use of eye movement,'' \emph{IEEE Intelligent Systems}, vol.~31,
  no.~3, pp. 49--56, 2016.

\bibitem{fridman2016owl}
L.~Fridman, J.~Lee, B.~Reimer, and T.~Victor, ``‘owl’and ‘lizard’:
  patterns of head pose and eye pose in driver gaze classification,'' \emph{IET
  Computer Vision}, vol.~10, no.~4, pp. 308--314, 2016.

\bibitem{vora2017generalizing}
S.~Vora, A.~Rangesh, and M.~M. Trivedi, ``On generalizing driver gaze zone
  estimation using convolutional neural networks,'' in \emph{Intelligent
  Vehicles Symposium (IV), 2017 IEEE}.\hskip 1em plus 0.5em minus 0.4em\relax
  IEEE, 2017, pp. 849--854.

\bibitem{vora2018driver}
------, ``Driver gaze zone estimation using convolutional neural networks: A
  general framework and ablative analysis,'' \emph{IEEE Transactions on
  Intelligent Vehicles}, vol.~3, no.~3, pp. 254--265, 2018.

\bibitem{Naqvi_NIR_2018}
R.~A. Naqvi, M.~Arsalan, G.~Batchuluun, H.~S. Yoon, and K.~R. Park, ``Deep
  learning-based gaze detection system for automobile drivers using a nir
  camera sensor,'' \emph{Sensors}, vol.~18, no.~2, p. 456, 2018.

\bibitem{Jha_2018}
S.~{Jha} and C.~{Busso}, ``Probabilistic estimation of the gaze region of the
  driver using dense classification,'' in \emph{2018 21st International
  Conference on Intelligent Transportation Systems (ITSC)}, Nov 2018, pp.
  697--702.

\bibitem{rangesh2020driver}
A.~Rangesh, B.~Zhang, and M.~M. Trivedi, ``Driver gaze estimation in the real
  world: Overcoming the eyeglass challenge,'' in \emph{2020 IEEE Intelligent
  Vehicles Symposium (IV)}.\hskip 1em plus 0.5em minus 0.4em\relax IEEE, 2020,
  pp. 1054--1059.

\bibitem{tawari2014continuous}
A.~Tawari, S.~Martin, and M.~M. Trivedi, ``Continuous head movement estimator
  for driver assistance: Issues, algorithms, and on-road evaluations,''
  \emph{IEEE Transactions on Intelligent Transportation Systems}, vol.~15,
  no.~2, pp. 818--830, 2014.

\bibitem{doshi2012head}
A.~Doshi and M.~M. Trivedi, ``Head and eye gaze dynamics during visual
  attention shifts in complex environments,'' \emph{Journal of vision},
  vol.~12, no.~2, pp. 9--9, 2012.

\bibitem{das2015performance}
N.~Das, E.~Ohn-Bar, and M.~M. Trivedi, ``On performance evaluation of driver
  hand detection algorithms: Challenges, dataset, and metrics,'' in \emph{2015
  IEEE 18th international conference on intelligent transportation
  systems}.\hskip 1em plus 0.5em minus 0.4em\relax IEEE, 2015, pp. 2953--2958.

\bibitem{ohn2013vehicle}
E.~Ohn-Bar and M.~Trivedi, ``In-vehicle hand activity recognition using
  integration of regions,'' in \emph{Intelligent Vehicles Symposium (IV), 2013
  IEEE}.\hskip 1em plus 0.5em minus 0.4em\relax IEEE, 2013, pp. 1034--1039.

\bibitem{ohn2014beyond}
E.~Ohn-Bar and M.~M. Trivedi, ``Beyond just keeping hands on the wheel: Towards
  visual interpretation of driver hand motion patterns,'' in \emph{Intelligent
  Transportation Systems (ITSC), 2014 IEEE 17th International Conference
  on}.\hskip 1em plus 0.5em minus 0.4em\relax IEEE, 2014, pp. 1245--1250.

\bibitem{ohn2014hand}
------, ``Hand gesture recognition in real time for automotive interfaces: A
  multimodal vision-based approach and evaluations,'' \emph{IEEE transactions
  on intelligent transportation systems}, vol.~15, no.~6, pp. 2368--2377, 2014.

\bibitem{borghi2018hands}
G.~Borghi, E.~Frigieri, R.~Vezzani, and R.~Cucchiara, ``Hands on the wheel: a
  dataset for driver hand detection and tracking,'' in \emph{Automatic Face \&
  Gesture Recognition (FG 2018), 2018 13th IEEE International Conference
  on}.\hskip 1em plus 0.5em minus 0.4em\relax IEEE, 2018, pp. 564--570.

\bibitem{rangesh2016hidden}
A.~Rangesh, E.~Ohn-Bar, and M.~M. Trivedi, ``Hidden hands: Tracking hands with
  an occlusion aware tracker,'' in \emph{Proceedings of the IEEE Conference on
  Computer Vision and Pattern Recognition Workshops}, 2016, pp. 19--26.

\bibitem{deo2016vehicle}
N.~Deo, A.~Rangesh, and M.~Trivedi, ``In-vehicle hand gesture recognition using
  hidden markov models,'' in \emph{Intelligent Transportation Systems (ITSC),
  2016 IEEE 19th International Conference on}.\hskip 1em plus 0.5em minus
  0.4em\relax IEEE, 2016, pp. 2179--2184.

\bibitem{molchanov2015hand}
P.~Molchanov, S.~Gupta, K.~Kim, and J.~Kautz, ``Hand gesture recognition with
  3d convolutional neural networks,'' in \emph{Proceedings of the IEEE
  conference on computer vision and pattern recognition workshops}, 2015, pp.
  1--7.

\bibitem{yuen2019looking}
K.~Yuen and M.~M. Trivedi, ``Looking at hands in autonomous vehicles: A convnet
  approach using part affinity fields,'' \emph{IEEE Transactions on Intelligent
  Vehicles}, vol.~5, no.~3, pp. 361--371, 2019.

\bibitem{rangesh2018handynet}
A.~Rangesh and M.~M. Trivedi, ``Handynet: A one-stop solution to detect,
  segment, localize \& analyze driver hands,'' \emph{arXiv preprint
  arXiv:1804.07834}, 2018.

\bibitem{tran2011pedal}
C.~Tran, A.~Doshi, and M.~M. Trivedi, ``Pedal error prediction by driver foot
  gesture analysis: A vision-based inquiry,'' in \emph{Intelligent Vehicles
  Symposium (IV), 2011 IEEE}.\hskip 1em plus 0.5em minus 0.4em\relax IEEE,
  2011, pp. 577--582.

\bibitem{tran2012modeling}
------, ``Modeling and prediction of driver behavior by foot gesture
  analysis,'' \emph{Computer Vision and Image Understanding}, vol. 116, no.~3,
  pp. 435--445, 2012.

\bibitem{rangesh2019foot}
A.~Rangesh and M.~Trivedi, ``Forced spatial attention for driver foot activity
  classification,'' in \emph{Proceedings of the IEEE International Conference
  on Computer Vision Workshops}, 2019, pp. 0--0.

\bibitem{rangesh2019forced}
------, ``Forced spatial attention for driver foot activity classification,''
  in \emph{Proceedings of the IEEE/CVF International Conference on Computer
  Vision Workshops}, 2019, pp. 0--0.

\bibitem{ohn2014head}
E.~Ohn-Bar, S.~Martin, A.~Tawari, and M.~M. Trivedi, ``Head, eye, and hand
  patterns for driver activity recognition,'' in \emph{Pattern Recognition
  (ICPR), 2014 22nd International Conference on}.\hskip 1em plus 0.5em minus
  0.4em\relax IEEE, 2014, pp. 660--665.

\bibitem{braunagel2015driver}
C.~Braunagel, E.~Kasneci, W.~Stolzmann, and W.~Rosenstiel, ``Driver-activity
  recognition in the context of conditionally autonomous driving,'' in
  \emph{Intelligent Transportation Systems (ITSC), 2015 IEEE 18th International
  Conference on}.\hskip 1em plus 0.5em minus 0.4em\relax IEEE, 2015, pp.
  1652--1657.

\bibitem{behera2018context}
A.~Behera, A.~Keidel, and B.~Debnath, ``Context-driven multi-stream lstm
  (m-lstm) for recognizing fine-grained activity of drivers,'' in \emph{Pattern
  Recognition}, 2019, pp. 298--314.

\bibitem{roitberg2020cnn}
A.~Roitberg, M.~Haurilet, S.~Rei{\ss}, and R.~Stiefelhagen, ``Cnn-based driver
  activity understanding: Shedding light on deep spatiotemporal
  representations,'' in \emph{2020 IEEE 23rd International Conference on
  Intelligent Transportation Systems (ITSC)}.\hskip 1em plus 0.5em minus
  0.4em\relax IEEE, 2020, pp. 1--6.

\bibitem{roitberg2020open}
A.~Roitberg, C.~Ma, M.~Haurilet, and R.~Stiefelhagen, ``Open set driver
  activity recognition,'' in \emph{Intelligent Vehicles Symposium (IV)}.\hskip
  1em plus 0.5em minus 0.4em\relax IEEE, 2020.

\bibitem{martin2019drive}
M.~Martin, A.~Roitberg, M.~Haurilet, M.~Horne, S.~Rei{\ss}, M.~Voit, and
  R.~Stiefelhagen, ``Drive\&act: A multi-modal dataset for fine-grained driver
  behavior recognition in autonomous vehicles,'' in \emph{Proceedings of the
  IEEE/CVF International Conference on Computer Vision}, 2019, pp. 2801--2810.

\bibitem{reiss2020deep}
S.~Rei{\ss}, A.~Roitberg, M.~Haurilet, and R.~Stiefelhagen, ``Deep
  classification-driven domain adaptation for cross-modal driver behavior
  recognition,'' in \emph{2020 IEEE Intelligent Vehicles Symposium (IV)}.\hskip
  1em plus 0.5em minus 0.4em\relax IEEE, 2020, pp. 1042--1047.

\bibitem{behera2020deep}
A.~Behera, Z.~Wharton, A.~Keidel, and B.~Debnath, ``Deep cnn, body pose and
  body-object interaction features for drivers' activity monitoring,''
  \emph{IEEE Transactions on Intelligent Transportation Systems}, 2020.

\bibitem{jain2015car}
A.~Jain, H.~S. Koppula, B.~Raghavan, S.~Soh, and A.~Saxena, ``Car that knows
  before you do: Anticipating maneuvers via learning temporal driving models,''
  in \emph{Proceedings of the IEEE International Conference on Computer
  Vision}, 2015, pp. 3182--3190.

\bibitem{jain2016recurrent}
A.~Jain, A.~Singh, H.~S. Koppula, S.~Soh, and A.~Saxena, ``Recurrent neural
  networks for driver activity anticipation via sensory-fusion architecture,''
  in \emph{Robotics and Automation (ICRA), 2016 IEEE International Conference
  on}.\hskip 1em plus 0.5em minus 0.4em\relax IEEE, 2016, pp. 3118--3125.

\bibitem{martin2018dynamics}
S.~Martin, S.~Vora, K.~Yuen, and M.~M. Trivedi, ``Dynamics of driver's gaze:
  Explorations in behavior modeling \& maneuver prediction,'' 2018.

\bibitem{ohn2014predicting}
E.~Ohn-Bar, A.~Tawari, S.~Martin, and M.~M. Trivedi, ``Predicting driver
  maneuvers by learning holistic features,'' in \emph{Intelligent Vehicles
  Symposium Proceedings, 2014 IEEE}.\hskip 1em plus 0.5em minus 0.4em\relax
  IEEE, 2014, pp. 719--724.

\bibitem{doshi2008comparative}
A.~Doshi and M.~Trivedi, ``A comparative exploration of eye gaze and head
  motion cues for lane change intent prediction,'' in \emph{2008 IEEE
  Intelligent Vehicles Symposium}.\hskip 1em plus 0.5em minus 0.4em\relax IEEE,
  2008, pp. 49--54.

\bibitem{shia2014semiautonomous}
V.~A. Shia, Y.~Gao, R.~Vasudevan, K.~D. Campbell, T.~Lin, F.~Borrelli, and
  R.~Bajcsy, ``Semiautonomous vehicular control using driver modeling,''
  \emph{IEEE Transactions on Intelligent Transportation Systems}, vol.~15,
  no.~6, pp. 2696--2709, 2014.

\bibitem{driggs2015improved}
K.~Driggs-Campbell, V.~Shia, and R.~Bajcsy, ``Improved driver modeling for
  human-in-the-loop vehicular control,'' in \emph{2015 IEEE International
  Conference on Robotics and Automation (ICRA)}.\hskip 1em plus 0.5em minus
  0.4em\relax IEEE, 2015, pp. 1654--1661.

\bibitem{liu2016driver}
T.~Liu, Y.~Yang, G.-B. Huang, Y.~K. Yeo, and Z.~Lin, ``Driver distraction
  detection using semi-supervised machine learning,'' \emph{IEEE transactions
  on intelligent transportation systems}, vol.~17, no.~4, pp. 1108--1120, 2016.

\bibitem{liang2007real}
Y.~Liang, M.~L. Reyes, and J.~D. Lee, ``Real-time detection of driver cognitive
  distraction using support vector machines,'' \emph{IEEE transactions on
  intelligent transportation systems}, vol.~8, no.~2, pp. 340--350, 2007.

\bibitem{liang2014hybrid}
Y.~Liang and J.~D. Lee, ``A hybrid bayesian network approach to detect driver
  cognitive distraction,'' \emph{Transportation research part C: emerging
  technologies}, vol.~38, pp. 146--155, 2014.

\bibitem{bergasa2006real}
L.~M. Bergasa, J.~Nuevo, M.~A. Sotelo, R.~Barea, and M.~E. Lopez, ``Real-time
  system for monitoring driver vigilance,'' \emph{IEEE Transactions on
  Intelligent Transportation Systems}, vol.~7, no.~1, pp. 63--77, 2006.

\bibitem{li2015predicting}
N.~Li and C.~Busso, ``Predicting perceived visual and cognitive distractions of
  drivers with multimodal features,'' \emph{IEEE Transactions on Intelligent
  Transportation Systems}, vol.~16, no.~1, pp. 51--65, 2015.

\bibitem{wollmer2011online}
M.~Wollmer, C.~Blaschke, T.~Schindl, B.~Schuller, B.~Farber, S.~Mayer, and
  B.~Trefflich, ``Online driver distraction detection using long short-term
  memory,'' \emph{IEEE Transactions on Intelligent Transportation Systems},
  vol.~12, no.~2, pp. 574--582, 2011.

\bibitem{gold2013take}
C.~Gold, D.~Damb{\"o}ck, L.~Lorenz, and K.~Bengler, ``“take over!” how long
  does it take to get the driver back into the loop?'' in \emph{Proceedings of
  the Human Factors and Ergonomics Society Annual Meeting}, vol.~57,
  no.~1.\hskip 1em plus 0.5em minus 0.4em\relax SAGE Publications Sage CA: Los
  Angeles, CA, 2013, pp. 1938--1942.

\bibitem{mok2015timing}
B.~K.-J. Mok, M.~Johns, K.~J. Lee, H.~P. Ive, D.~Miller, and W.~Ju, ``Timing of
  unstructured transitions of control in automated driving,'' in \emph{2015
  IEEE intelligent vehicles symposium (IV)}.\hskip 1em plus 0.5em minus
  0.4em\relax IEEE, 2015, pp. 1167--1172.

\bibitem{radlmayr2014traffic}
J.~Radlmayr, C.~Gold, L.~Lorenz, M.~Farid, and K.~Bengler, ``How traffic
  situations and non-driving related tasks affect the take-over quality in
  highly automated driving,'' in \emph{Proceedings of the human factors and
  ergonomics society annual meeting}, vol.~58, no.~1.\hskip 1em plus 0.5em
  minus 0.4em\relax Sage Publications Sage CA: Los Angeles, CA, 2014, pp.
  2063--2067.

\bibitem{gold2016taking}
C.~Gold, M.~K{\"o}rber, D.~Lechner, and K.~Bengler, ``Taking over control from
  highly automated vehicles in complex traffic situations: the role of traffic
  density,'' \emph{Human factors}, vol.~58, no.~4, pp. 642--652, 2016.

\bibitem{korber2016influence}
M.~K{\"o}rber, C.~Gold, D.~Lechner, and K.~Bengler, ``The influence of age on
  the take-over of vehicle control in highly automated driving,''
  \emph{Transportation research part F: traffic psychology and behaviour},
  vol.~39, pp. 19--32, 2016.

\bibitem{clark2017age}
H.~Clark and J.~Feng, ``Age differences in the takeover of vehicle control and
  engagement in non-driving-related activities in simulated driving with
  conditional automation,'' \emph{Accident Analysis \& Prevention}, vol. 106,
  pp. 468--479, 2017.

\bibitem{petermeijer2017take}
S.~Petermeijer, P.~Bazilinskyy, K.~Bengler, and J.~De~Winter, ``Take-over
  again: Investigating multimodal and directional tors to get the driver back
  into the loop,'' \emph{Applied ergonomics}, vol.~62, pp. 204--215, 2017.

\bibitem{huang2019multimodal}
G.~Huang, C.~Steele, X.~Zhang, and B.~J. Pitts, ``Multimodal cue combinations:
  a possible approach to designing in-vehicle takeover requests for
  semi-autonomous driving,'' in \emph{Proceedings of the Human Factors and
  Ergonomics Society Annual Meeting}, vol.~63, no.~1.\hskip 1em plus 0.5em
  minus 0.4em\relax SAGE Publications Sage CA: Los Angeles, CA, 2019, pp.
  1739--1743.

\bibitem{dogan2017transition}
E.~Dogan, M.-C. Rahal, R.~Deborne, P.~Delhomme, A.~Kemeny, and J.~Perrin,
  ``Transition of control in a partially automated vehicle: Effects of
  anticipation and non-driving-related task involvement,'' \emph{Transportation
  research part F: traffic psychology and behaviour}, vol.~46, pp. 205--215,
  2017.

\bibitem{eriksson2017takeover}
A.~Eriksson and N.~A. Stanton, ``Takeover time in highly automated vehicles:
  noncritical transitions to and from manual control,'' \emph{Human factors},
  vol.~59, no.~4, pp. 689--705, 2017.

\bibitem{naujoks2019noncritical}
F.~Naujoks, C.~Purucker, K.~Wiedemann, and C.~Marberger, ``Noncritical state
  transitions during conditionally automated driving on german freeways:
  Effects of non--driving related tasks on takeover time and takeover
  quality,'' \emph{Human factors}, vol.~61, no.~4, pp. 596--613, 2019.

\bibitem{braunagel2017ready}
C.~Braunagel, W.~Rosenstiel, and E.~Kasneci, ``Ready for take-over? a new
  driver assistance system for an automated classification of driver take-over
  readiness,'' \emph{IEEE Intelligent Transportation Systems Magazine}, vol.~9,
  no.~4, pp. 10--22, 2017.

\bibitem{lotz2018predicting}
A.~Lotz and S.~Weissenberger, ``Predicting take-over times of truck drivers in
  conditional autonomous driving,'' in \emph{International Conference on
  Applied Human Factors and Ergonomics}.\hskip 1em plus 0.5em minus 0.4em\relax
  Springer, 2018, pp. 329--338.

\bibitem{berghofer2019prediction}
F.~Bergh{\"o}fer, C.~Purucker, F.~Naujoks, K.~Wiedemann, and C.~Marberger,
  ``Prediction of take-over time demand in highly automated driving. results of
  a naturalistic driving study prediction of take-over time demand in
  conditionally automated driving-results of a real world driving study,''
  \emph{Proceedings of the Human Factors and Ergonomics Society Europe}, 2019.

\bibitem{du2020predicting}
N.~Du, F.~Zhou, E.~M. Pulver, D.~M. Tilbury, L.~P. Robert, A.~K. Pradhan, and
  X.~J. Yang, ``Predicting driver takeover performance in conditionally
  automated driving,'' \emph{Accident Analysis \& Prevention}, vol. 148, p.
  105748, 2020.

\bibitem{hwang2020predicting}
S.~Hwang, A.~G. Banerjee, and L.~N. Boyle, ``Predicting driver's transition
  time to a secondary task given an in-vehicle alert,'' \emph{IEEE Transactions
  on Intelligent Transportation Systems}, 2020.

\bibitem{pakdamanian2021deeptake}
E.~Pakdamanian, S.~Sheng, S.~Baee, S.~Heo, S.~Kraus, and L.~Feng, ``Deeptake:
  Prediction of driver takeover behavior using multimodal data,'' in
  \emph{Proceedings of the 2021 CHI Conference on Human Factors in Computing
  Systems}, 2021, pp. 1--14.

\bibitem{yuen2016looking}
K.~Yuen, S.~Martin, and M.~M. Trivedi, ``Looking at faces in a vehicle: A deep
  cnn based approach and evaluation,'' in \emph{Intelligent Transportation
  Systems (ITSC), 2016 IEEE 19th International Conference on}.\hskip 1em plus
  0.5em minus 0.4em\relax IEEE, 2016, pp. 649--654.

\bibitem{yuen2018looking}
K.~Yuen and M.~M. Trivedi, ``Looking at hands in autonomous vehicles: A convnet
  approach using part affinity fields,'' \emph{arXiv preprint
  arXiv:1804.01176}, 2018.

\bibitem{cao2017realtime}
Z.~Cao, T.~Simon, S.-E. Wei, and Y.~Sheikh, ``Realtime multi-person 2d pose
  estimation using part affinity fields,'' in \emph{CVPR}, 2017.

\end{thebibliography}

\end{document}